\documentclass{article}

\usepackage[preprint]{neurips_2024}

\usepackage[utf8]{inputenc} 
\usepackage[T1]{fontenc}    
\usepackage{hyperref}       
\usepackage{url}            
\usepackage{booktabs}       
\usepackage{amsfonts}       
\usepackage{nicefrac}       
\usepackage{microtype}      
\usepackage{xcolor}         
\usepackage{tikz}
\usepackage{amsmath}
\usepackage{algorithm2e}
\usepackage{multirow}
\usepackage{siunitx}
\usetikzlibrary{calc, positioning, arrows.meta, automata}

\newcommand{\dox}[0]{\text{do}}

\title{Controlling for Unobserved Confounding with Large Language Model Classification of Patient Smoking Status}

\author{%
  Samuel Lee \\
  Department of Computer Science\\
  Northwestern University\\
  Evanston, IL 60201 \\
  \texttt{samlee2023@u.northwestern.edu}
  \And 
  Zach Wood-Doughty \\
  Department of Computer Science\\
  Northwestern University\\
  Evanston, IL 60201 \\
  \texttt{zach@northwestern.edu} \\
}

\begin{document}

\maketitle

\begin{abstract}

Causal understanding is a fundamental goal of evidence-based medicine.
When randomization is impossible, causal inference methods allow the estimation of treatment effects from retrospective analysis of observational data.
However, such analyses rely on a number of assumptions, often including that of no unobserved confounding.
In many practical settings, this assumption is violated when important variables are not explicitly measured in the clinical record.
Prior work has proposed to address unobserved confounding with machine learning by imputing unobserved variables and then correcting for the classifier's mismeasurement.
When such a classifier can be trained and the necessary assumptions are met, this method can recover an unbiased estimate of a causal effect. 
However, such work has been limited to synthetic data, simple classifiers, and binary variables.
This paper extends this methodology by using a large language model trained on clinical notes to predict patients' smoking status, which would otherwise be unobserved.
We then apply a measurement error correction on the categorical predicted smoking status to estimate the causal effect of transthoracic echocardiography on mortality in the MIMIC dataset. 
  
\end{abstract}



\section{Introduction}
\label{sec:intro}

Evidence-based medicine seeks to use data from clinical research to inform best practices \citep{sackett1996evidence,masic2008evidence}.
While randomized control trials (RCTs) are the gold standard of clinical research, randomization is often impossible or unethical \citep{sanson2007limitations}.
In the absence of randomization, retrospective causal analyses
must control for confounding variables that could influence both a patient's treatment assignment and their outcome \citep{vanderweele2013definition}.
For example, if a certain medication is only given to the patients with the highest risk, a naive correlational analysis may suggest that the medication itself increases the risk of mortality.
Causal inference methods establish causal claims by controlling for confounding and other sources of bias \citep{pearl2009causality}.

All causal inference methods rely on assumptions about the underlying data-generating process; a common but untestable assumption is the absence of unobserved confounding \citep{groenwold2009quantitative}.
If a confounding variable is unobserved, it is in general impossible to estimate the desired causal effect without bias \citep{shpitser2006identification}.
Even with the wide availability of electronic health record (EHR) data, it may be difficult to rule out the possibility of unobserved confounding \citep{groenwold2008quantifying}.
Many causal methods attempt to detect and mitigate unobserved confounding \citep{flanders2011method,baiocchi2014instrumental,uddin2016methods,tchetgen2020introduction}.

One such methodological approach considers the case where a necessary confounder is unobserved, but a proxy for that variable is instead observed.
A canonical example is Medicaid enrollment as a proxy for socioeconomic status (SES), because Medicaid eligibility is tied to household income \citep{simpson2020implications}.
While a naive approach of treating SES and Medicaid enrollment as equivalent variables could introduce bias \citep{greenland1985confounding,ogburn2012nondifferential},
methods developed by \citet{pearl2010measurement} and \citet{kuroki2014measurement} can recover unbiased causal estimates through a measurement error correction.
When such a proxy (e.g., Medicaid enrollment) is unavailable, one option may be to train a machine learning (ML) classifier to predict the values of the unobserved confounder and then apply this measurement error correction to account for the classifier's errors \citep{wooddoughty2020sensitivity}.

This paper builds upon this line of work, extending prior methods from simple classifiers and synthetic data to large language models (LLMs) and two real-world datasets.
Following \citet{feng2018tte}, we analyze the causal effect of transthoracic echocardiography (TTE) on 28-day mortality among ICU patients diagnosed with sepsis.
A limitation of prior work is its inability to account for patient smoking status; this variable is not explicitly recorded in the EHR data, but tobacco use is considered a risk factor for mortality among patients with sepsis \citep{alroumi2018impact,zhang2022association}.
Smoking status could also plausibly affect a clinician's decision to administer TTE, which would make it an unobserved confounder.
We train and apply an ML classifier to predict patients' smoking status and estimate the model's error rate \citep{mulyar2019pheno}.
We extend the measurement error correction to categorical variables and use it to debias our causal estimates.
Our method produces estimates with modest variance and our results provide evidence that TTE prevents mortality, with or without accounting for patient smoking status. 

This primary contribution of this paper is a methodology for imputing unmeasured variables with large language models and then estimating unbiased causal effects with a measurement error correction. We release our code to enable future research.\footnote{\url{https://github.com/controllingunobservedpaper/NEURIPS_submission_2024}}

\section{Related Work}
\label{sec: related_works}

\subsection{ML and LLMs for Clinical Prediction}

Machine learning (ML) methods have advanced over the past decade to provide impressive real-world performance in a wide variety of domains.
For text data, large language models (LLMs) have become the dominant paradigm for text generation and classification tasks \citep{zhou2023survey}.
The most common approach for training LLMs uses generative pretraining followed by supervised task-specific fine-tuning \citep{radford2018improving}.
In the clinical domain, LLMs have been applied to a wide variety of tasks, such as 30-day all-cause readmission prediction, in-hospital mortality prediction, comorbidity index prediction, length of stay prediction, and insurance denial prediction \citep{jiang2023health}.
While such methods can demonstrate state-of-the-art performance on a variety of datasets, their applications to real-world clinical practice have been somewhat limited \citep{chen2017machine}.
Such models are complicated and often uninterpretable, which may prevent clinicians from trusting the reasoning underlying their predictions \citep{tonekaboni2019clinicians}.
Additionally, test set evaluations may not be fully indicative of real-world performance in the presence of domain shift and other sources of bias \citep{subbaswamy2020development,finlayson2021clinician}.

We focus on the specific task of predicting patient smoking status from clinical notes, using a labeled dataset released by \citet{uzuner2008n2c2}.
Variations on this task have been a sustained focus of ML applications because of the relevance of tobacco use to a wide variety of health outcomes \citep{sohn2009mayo,palmer2019building}.
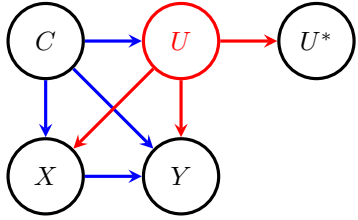
\begin{figure}
\begin{center}
\begin{tikzpicture}[>=stealth, node distance=1.8cm]
    \tikzstyle{obs} = [draw, very thick, circle, minimum size=1cm, inner sep=1.8pt]

    \begin{scope}
        \path[->, very thick]
            node[obs] (C) {$C$}
            node[obs, red, right of=C] (U) {$U$}
            node[obs, right of=U] (UU) {$U^*$}
            node[obs, below of=C] (X) {$X$}
            node[obs, right of=X] (Y) {$Y$}

            (C) edge[blue] (X)
            (C) edge[blue] (Y)
            (U) edge[red] (X)
            (U) edge[red] (Y)
            (X) edge[blue] (Y)
            (C) edge[blue] (U)
            (U) edge[red] (UU)
        ;
    \end{scope}
\end{tikzpicture}
\caption{DAG with treatment $X$, outcome $Y$, and observed confounders $C$. $U^*$ is the observed but noisy proxy for the unobserved confounder $U$.}
\label{fig:DAG}
\end{center}
\end{figure}

\subsection{Causal Inference}

Whereas practical applications of ML methods to the clinical domain are a relatively recent phenomenon \citep{jiang2023health},
many of the foundational methods in causal inference were developed to enable epidemiological studies \citep{pearl2009causality}.
In this work, we use the potential outcomes framework that explores the causal effect of a treatment $X$ on an outcome $Y$:
specifically, following \citet{feng2018tte}, the causal effect of TTE on 28-day mortality among patients with sepsis.
We use causal do-notation $p(Y \mid \dox(x))$ to denote the distribution of a patient's hypothetical mortality had they be
\emph{randomly assigned} TTE value $x$ \citep{pearl2012calculus}.
In an RCT, $E[Y \mid X=x]$ is equivalent to $E[Y \mid \dox(x)]$ because correlation between the randomly-assigned treatment and outcome implies (in probability) a causal effect \citep{pearl2009causality}.
Because in our dataset TTE was not randomly assigned, we need additional assumptions to \emph{identify} the counterfactual distribution $E[Y \mid \dox(x)]$ from the observed data.

We use directed acyclic graphs (DAGs) to represent our assumptions about the data \citep{pearl1995causal}.
Figure \ref{fig:DAG} shows the causal DAG for our problem.
In addition to our treatment $X$ and outcome $Y$, $C$ represents our vector of observed confounders, including patient age, gender, and several clinical markers. $U$ is smoking status, which is hypothesized as a possible confounder but is unobserved in our data. $U^*$, introduced in \S \ref{subsec:related_measurement_error}, is our noisy proxy.

Both our treatment and outcomes are binary variables, and our causal estimands of interest are the risk ratio and odds ratio, equations (\ref{eq:rr_defn}) and (\ref{eq:or_defn}) below, respectively \citep{hernan2020causal}:
\begingroup
\thinmuskip=1mu plus 1mu
\medmuskip=3mu plus 2mu minus 4mu
\thickmuskip=3mu plus 3mu
\begin{align}
    \text{RR}&: \dfrac{p(Y=1 \mid \dox(x=1))}{p(Y=1 \mid \dox(x=0))}
    \label{eq:rr_defn} \\[1em]
    \text{OR}&: \dfrac{p(Y=1 \mid \dox(x=1)) p(Y=0 \mid \dox(x=0))}{p(Y=0 \mid \dox(x=1)) p(Y=1 \mid \dox(x=0))}
    \label{eq:or_defn}
\end{align}
\endgroup

\subsection{Measurement Error}
\label{subsec:related_measurement_error}

If all confounders $C$ and $U$ were observed, then $p(Y \mid \dox(x))$ is identified -- as we show in Equation \ref{eq:simple_do3} below -- and can be estimated in a number of ways \citep{glynn2010introduction}.
If $U$ is unobserved, however, then the causal effect of $X$ on $Y$ is \emph{unidentified} and the problem becomes impossible without additional assumptions \citep{shpitser2006identification}.
We assume access to $U^*$, a noisy proxy with the same cardinality as $U$ that is governed by some noisy relationship $p(U^* \mid U)$.
If a matrix representation of this probability $p(U^* \mid U)$ can be inverted, then the work of \citet{pearl2010measurement} and \citet{kuroki2014measurement} show that we can recover $p(Y \mid \dox(x))$.
This method is known as and alternatively referred to as `matrix adjustment' or `effect restoration' \citep{kuroki2014measurement}; we demonstrate this derivation in \S\ref{subsec:methods_me}.
Throughout this paper, following \S \ref{subsec:methods_clf}, we assume that $U^*$ is predicted by an LLM classifier and that $p(U^* \mid U)$ is the error rate of the classification.
Our method assumes no \emph{additional} unobserved confounding other than $U$.

An alternative to matrix adjustment methods is simulation-extrapolation (SIMEX), a widely-used method for measurement error \citep{cook1994simex}.
SIMEX generally requires strong parametric assumptions about the data \citep{sevilimedu2022simex}, prior knowledge of the known measurement error variance \citep{cook1994simex}, and the unknown confounder to be continuous \citep{lederer2006simex}.
To overcome these limitations, \citet{kuch2006mcsimex} proposed MC-SIMEX which avoids parametric assumptions but has weaker theoretical foundations.
While we focus on developing matrix correction methods and leave a comprehensive survey of SIMEX methods for other work \citep{sevilimedu2022simex},
we compare our proposed methodology against MC-SIMEX in our experiments.




\section{Methodology}
\label{sec: methodology}

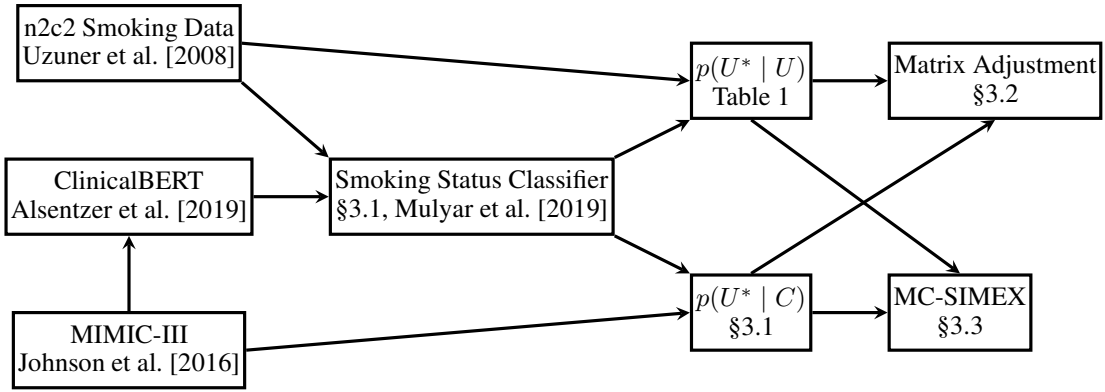
\begin{figure}
\begin{center}
\vspace{1em}
\begin{tikzpicture}[>=stealth, node distance=2cm]
    \tikzstyle{obs} = [draw, very thick, rectangle, minimum width=1cm, minimum height=1cm, align=center, inner sep=1.8pt]
    \tikzstyle{circ} = [draw, very thick, circle, minimum width=1cm, minimum height=1cm, align=center, inner sep=1.8pt]

    \begin{scope}

        
        \node[obs] (CurrBERT) {Smoking Status Classifier \\ \S\ref{subsec:methods_clf}, \citet{mulyar2019pheno}};

        \node[obs] (ClinicalBERT) [left = 1cm of CurrBERT]
        {ClinicalBERT \\ \citet{alsentzer2019publicly}};
        
        \node[obs] (MIMIC) [below =1 cm of ClinicalBERT] {MIMIC-III \\ \citet{johnson2016mimic}};
        
        \node[obs] (TrainingData) [above = 1cm of ClinicalBERT] {n2c2 Smoking Data \\ \citet{uzuner2008n2c2}};
        
        \node[obs] (Error) [above right = 0.5cm and 1cm of CurrBERT] {$p(U^* \mid U)$ \\ Table \ref{tab:confusion_matrix}};
        
        \node[obs] (Predictions) [below right = 0.5cm and 1cm of CurrBERT]{$p(U^* \mid C)$ \\ \S\ref{subsec:methods_clf}};
        
        \node[obs] (Matrix) [right = 1cm of Error] {Matrix Adjustment \\ \S\ref{subsec:methods_me}};
        
        \node[obs] (SIMEX) [right = 1cm of Predictions] {MC-SIMEX \\ \S\ref{subsec:mcsimex_corr}};

        \draw[->, very thick] (ClinicalBERT.east) -- (CurrBERT.west);
        
        \draw[->, very thick] (TrainingData.south east) -- (CurrBERT.north west);
        \draw[->, very thick] (TrainingData.east) -- (Error.west);

        \draw[->, very thick] (MIMIC.north) -- (ClinicalBERT.south);        
        \draw[->, very thick] (MIMIC.east) -- (Predictions.west);

        \draw[->, very thick] (CurrBERT.north east) -- (Error.south west);
        \draw[->, very thick] (CurrBERT.south east) -- (Predictions.north west);
        \draw[->, very thick] (Error.east) -- (Matrix.west);
        \draw[->, very thick] (Predictions.north) -- (Matrix.south);

        \draw[->, very thick] (Error.south) -- (SIMEX.north);
        \draw[->, very thick] (Predictions.east) -- (SIMEX.west);
        
    \end{scope}
\end{tikzpicture}
\caption{Overview of training and estimation methodology}
\label{fig:MethodologyProcess}
\end{center}
\end{figure}

We divide our methodological approach into two sections.
First, we discuss the classifier we use to predict smoking status for patients in the MIMIC dataset.
We discuss the model's training and validation procedure and its classification of our data.
Then, we incorporate our model's predictions and our estimate of its error rate into the measurement error formulation to estimate causal effects.
We discuss our extensions of past work to categorical mismeasured variables and high-dimensional observed confounders. Figure \ref{fig:MethodologyProcess} shows the high-level overview of our entire methodology, including the pretrained LLMs and most important data sources.

\subsection{Smoking Status Classifier}
\label{subsec:methods_clf}

Our modeling approach most closely follows that of \citet{mulyar2019pheno}.
We start with a ClinicalBERT LLM architecture using the weights that were released by \citet{alsentzer2019publicly} after pretraining on PubMed abstracts, PMC articles, and all MIMIC-III notes \citep{alsentzer2019publicly, lee2019biobert}.
We then adapt this LLM by training an LSTM on top of its representation to predict categorical smoking status in the 2006 n2c2 smoking dataset \citep{uzuner2008n2c2}.
The training dataset contains 398 clinical notes and corresponding labels of the patient as either past smoker, current smoker, non-smoker, and unknown.
These labels were manually annotated by pulmonologists.
We follow the hyperparameter choices of \citet{mulyar2019pheno} and train for 1,000 epochs. 

After training, we evaluate our model on the test set of 101 additional patient records.
Our model achieves 86.1\% accuracy with the confusion matrix shown in Table \ref{tab:confusion_matrix}.\footnote{As accuracy here is equivalent to micro-averaged F1, our model performs lower than the 92.8\% accuracy reported by \citet{mulyar2019pheno}. We hypothesize this is simply due to random variation or changes in the Pytorch libraries.}
This confusion matrix, after dividing by the row sums to produce a stochastic matrix, provides the error-rate matrix $p(U^* \mid U)$ that will be used in our measurement error correction in the following section.

To apply this model to patients in the MIMIC dataset, we first preprocess that data using the MIMIC-EXTRACT approach \citep{wang2020mimicextract}.
We then follow the preprocessing steps of \citet{feng2018tte} to enable a direct comparison against their analysis.
Our trained classifier is then used to predict smoking status for the patients in the evaluation set of MIMIC patients who met the criteria for sepsis \citep{angus2016framework}.
Of those, 2,058 were classified as past smoker, 93 as current smoker, 1,413 as non-smoker, and 1,171 as unknown.
We discard 64 patients for whom the classifier did not make a prediction due to a confidence threshold set by \citet{mulyar2019pheno}.
Specifically, the model makes no prediction if its final {\tt tanh} activation outputs a negative logit value for all four classes.

\begin{table}[t]
    \caption{Confusion matrix for model predictions of smoking status on the n2c2 test set \citep{uzuner2008n2c2}. True labels are on the rows, predicted values on the columns.}
    \label{tab:confusion_matrix}
    \centering
    \begin{tabular}{l c c c c}
        & Past & Current & Never & Unknown \\
        \midrule
        Past    & 8 & 0 &  2 & 1  \\
        Current & 4 & 4 &  3 & 0  \\
        Never   & 1 & 0 & 14 & 1  \\
        Unknown & 1 & 0 &  1 & 61 \\
        \bottomrule
    \end{tabular}
    
\end{table}

\subsection{Matrix Adjustment}
\label{subsec:methods_me}
We can now incorporate our smoking status predictions into our estimate of the causal effect of TTE on 28-day mortality.
Figure \ref{fig:DAG} portrays our assumptions about the underlying data distribution.
Our treatment $X$ is TTE, our outcome $Y$ is 28-day mortality, and $C$ is a vector of 39 observed covariates.\footnote{See Table 1 of \citet{feng2018tte} for a complete list.}
Patient smoking status $U$ is our unobserved confounder, and $U^*$ is our classifier's prediction.
Given this notation, we have data on the joint probability $p(X, Y, C, U^*)$ from patients in MIMIC.
We additionally have the error rate $P(U^* \mid U)$ from our classifiers validation on the n2c2 dataset.
We assume that the classifier's error rate remains constant between the two datasets; future work could relax this assumption by collecting additional ground-truth annotations on the MIMIC dataset.
We further make a \emph{non-differential error} assumption, that:
\begin{align}
    p(U^* \mid U, X, Y, C) &= P(U^* \mid U) \label{eq:nondiff}
\end{align}

Because our smoking status is a categorical variable with four values, our error rate distribution $p(U^* \mid U)$ can be represented as a 4x4 matrix $M(U^*, U)$ computed from the confusion matrix in Table \ref{tab:confusion_matrix} by dividing each cell by its corresponding row sum.
In general, using $u_i$ to denote $U=i$ and $u_j^*$ to denote $U^*=j$, this can be written as:
\begin{equation*}
\begin{bmatrix}
P(u_0^* \mid u_0) & P(u_0^* \mid u_1) & P(u_0^* \mid u_2) & P(u_0^* \mid u_3) \\
P(u_1^* \mid u_0) & P(u_1^* \mid u_1) & P(u_1^* \mid u_2) & P(u_1^* \mid u_3) \\
P(u_2^* \mid u_0) & P(u_2^* \mid u_1) & P(u_2^* \mid u_2) & P(u_2^* \mid u_3) \\
P(u_3^* \mid u_0) & P(u_3^* \mid u_1) & P(u_3^* \mid u_2) & P(u_3^* \mid u_3)
\end{bmatrix}
\end{equation*}

We can use $M(U^*, U)$ to write our observed data distribution as a function of the underlying distribution and our error rate $p(U^* \mid U)$.
\begin{align}
    p(Y, X, C, U^*)
    &\;= \sum_{U} p(U^* \mid Y, X, C, U) p(Y, X, C, U)  \label{eq:deriv3} \\
    &\;= \sum_{U} p(U^* \mid U) p(Y, X, C, U)  \label{eq:deriv4} \\
    &\;= M(U^*, U) \times p(Y, X, C, U) \label{eq:deriv5}
\end{align}

Equation \ref{eq:deriv3} holds by rules of probability, (\ref{eq:deriv4}) holds by (\ref{eq:nondiff}) above, and (\ref{eq:deriv5}) is rewritten to formulate this summation over probability distributions as a matrix multiplication.
This follows the approach of \citet{pearl2010measurement}; for fixed values of $Y, X, C$, $p(Y, X, C, U)$ is a vector of four values.
We then multiply that vector by our 4x4 error matrix $M(U^*, U)$ to get a different vector of four values.

Assuming $M(U^*, U)$ has an inverse $I(U^*, U)$, we then have:
\vspace{0.3em}
\begin{align}
    p(Y, X, C, U) = I(U^*, U) \times p(Y, X, C, U^*) \label{eq:deriv6}    
\end{align}

This inverse matrix $I(U^*, U)$ allows us, given $p(Y, X, C, U^*)$ and $M(U^*, U)$, to recover $p(Y, X, C, U)$.
This derivation follows those presented in prior work by \citet{pearl2010measurement} and \citet{kuroki2014measurement}.
To move from $p(Y, X, C, U)$ to our counterfactual $p(Y \mid \dox(x))$, we can follow a standard do-calculus derivation:

\begin{align}
    p(Y \mid \dox(x))
    &= \sum_{C, U} p(Y \mid \dox(x), C, U)p(C, U \mid \dox(x)) \label{eq:simple_do1} \\
    &= \sum_{C, U} p(Y \mid X=x, C, U)p(C, U \mid \dox(x)) \label{eq:simple_do2} \\
    &= \sum_{C, U} p(Y \mid X=x, C, U)p(C, U) \label{eq:simple_do3}
\end{align}
Equation \ref{eq:simple_do1} hold by marginalization and chain rule, (\ref{eq:simple_do2}) holds by Rule 2 of do-calculus, and (\ref{eq:simple_do3}) holds by Rule 3 of do-calculus \citep{pearl2012calculus}.\footnote{For a gentle and thorough derivation of this, see \url{https://www.andrewheiss.com/blog/2021/09/07/do-calculus-backdoors/}.}
However, the relative simplicity of the proof that the counterfactual is \emph{identified} obscures some of the practical challenges of \emph{estimating} causal effects in our specific domain application.
To implement estimators for the risk and odds ratios requires fitting models to these probability distributions.
Unlike for example in \citet{wooddoughty2020sensitivity} where there is a single binary observed confounder, we have 39 covariates.

As Equation \ref{eq:simple_do3} is a function of $p(Y \mid X, C, U)$ and $p(C, U)$, we can first use marginalization and conditioning to derive these from $p(Y, X, C, U)$ as provided to us by Equation \ref{eq:deriv6}.
To actually compute Equation \ref{eq:deriv6}, we fit three models: logistic regressions for $p(Y\mid X, C, U^*)$ and $p(X \mid C)$, and a multinomial logit model $p(U^* \mid X, C)$.\footnote{Note that replacing $p(Y \mid \cdot)$ with a linear regression $E[Y \mid \cdot]$ would trivially extend this to applications with continuous outcomes where the risk difference is the estimand of choice.}
All models are fit using the \emph{predicted} smoking statuses (i.e., using $U^*$, not on $U$); we perform the measurement error correction \emph{after} fitting models.

\begingroup
\thinmuskip=1mu plus 1mu
\medmuskip=3mu plus 2mu minus 4mu
\thickmuskip=3mu plus 3mu
\begin{align}
    p(Y, X, U \mid C)
    &= I(U^*, U)p(Y, X, U^* \mid C) \nonumber \\[0.2em]
    &= I(U^*, U)p(Y \mid X, U^*, C)p(U^* \mid X, C)p(X \mid C) \label{eq:yxu_c}
\end{align}
\endgroup

Because $C$ is high-dimensional, we avoid modeling it and use the empirical approximation $1/N$. 

\begin{align}
    p(U \mid C) &=\sum_{X} p(U, X \mid C) \nonumber \\
    &= \sum_{X} I(U^*, U)p(U^*, X \mid C) \nonumber \\
    &= \sum_{X} I(U^*, U)p(U^* \mid X, C)p(X \mid C) \label{eq:uc_full} \\[0.2em]
    p(U, C) &\approx \frac{1}{N}\sum_{i=1}^N p(U \mid C_i) \label{eq:uc}
\end{align}

While Equation \ref{eq:yxu_c} defines $p(Y, X, U \mid C)$, we can easily transform it to calculate the conditional probability we need for our counterfactual.

\begin{align}
    &p(Y \mid X, C, U) = \dfrac{p(Y, X, U \mid C)}{\sum_y p(Y=y, X, U \mid C)} \label{eq:y_xuc}
\end{align}

Then, plugging Equations \ref{eq:y_xuc} and \ref{eq:uc} into (\ref{eq:simple_do3}), we have written our counterfactual $p(Y \mid \dox(x))$ as a function of our observed data distribution.
We can plug that counterfactual into the definitions of Equations \ref{eq:rr_defn} and \ref{eq:or_defn} to estimate our causal effects.

\begin{table}[]
    \caption{Comparison of our method's causal estimates against those of the five methods from Table 2 of \citet{feng2018tte}. DR: Doubly Robust; PS: Propensity Score; IPW: Inverse Propensity Weighting. All results indicate a protective effect of TTE against mortality.}
    \label{tab:or_results}
    \centering
    \begin{tabular}{l c c}
                        & Odds Ratio & Confidence Interval\\
        \midrule
        Matrix \S\ref{subsec:methods_me}       & 0.89      & [0.60, 0.98]  \\  
        MC-SIMEX \S\ref{subsec:mcsimex_corr}& 0.90 & [0.83, 1.00] \\
        \midrule
        DR Unbalanced   & 0.78      & [0.68, 0.90]        \\
        DR All          & 0.64      & [0.52, 0.78]        \\
        PS IPW          & 0.84      & [0.78, 0.92]        \\
        PS Matching     & 0.78      & [0.66, 0.92]        \\
        Multivariate    & 0.64      & [0.53, 0.78]        \\
        \bottomrule
    \end{tabular}
\end{table}

\begin{table}[!t]
    \caption{Boostrapped 95\% confidence intervals for Risk Ratio and Odds Ratio estimates. The `None' row shows the point estimate without bootstrapping. Each other row is computed across 100 total resampled datasets, with bootstrapping applied to the MIMIC dataset used to fit our models, the n2c2 test set used to estimate $M(U^*, U)$, or both datasets.}
    \label{tab:bootstrap}
    \centering
    \begin{tabular}{l ccccc}
        & \multicolumn{2}{c}{Matrix Adjustment} & &\multicolumn{2}{c}{MC-SIMEX} \\[0.1em]
        \cmidrule{2-3} \cmidrule{5-6}
         Bootstrap      & RR           & OR             && RR   & OR \\
        \midrule
        None            & 0.88         & 0.89           && 0.93         & 0.90         \\    
        MIMIC           & [0.85, 0.90] & [0.87, 0.90]   && [0.89, 0.96] & [0.76, 0.95] \\   
        n2c2            & [0.77, 0.99] & [0.77, 0.98]   && [0.93, 0.94] & [0.90, 0.91] \\
        Both            & [0.65, 0.92] & [0.60, 0.98]   && [0.85, 1.01] & [0.83, 1.00] \\         
        \bottomrule
    \end{tabular}
\end{table}

\subsection{MC-SIMEX}
\label{subsec:mcsimex_corr}

Following our discussion in \S\ref{subsec:related_measurement_error}, we compare our proposed matrix adjustment method against MC-SIMEX, another common method for handling measurement error.
While the original SIMEX method was designed for continuous unobserved variables, MC-SIMEX allows for categorical variables \citep{lederer2006simex}.
We use the R package released by the authors.\footnote{\url{https://github.com/cran/simex}}
As indicated in Figure \ref{fig:MethodologyProcess}, SIMEX also requires access to both the noisy imputed $U^*$ values and our estimate of the error rate $p(U^* \mid U)$.
Despite the similarities of these methods, our matrix adjustment approach has the benefit of more established theoretical foundations, and makes no assumptions about the parametric form of the estimator.
It can also be easily applied to more complicated counterfactual estimands where the identifying functional of the observed data is not simply a linear model of the treatment and covariates, e.g., a front-door or proximal estimator \citep{bellemare2020paper,cui2023semiparametric}.



\section{Results}
\label{sec:results}

Our experimental analysis follows Figure \ref{fig:MethodologyProcess} in applying both matrix adjustment and MC-SIMEX to the MIMIC-III dataset as preprocessed and analyzed by \citet{feng2018tte}.
We use our trained LLM classifier to impute patients' smoking status, fit our models for $p(Y \mid X, U^*, C)$, $p(U^* \mid X, C)$, and $p(X \mid C)$, and then incorporate those models and our estimated error rates $p(U^* \mid U)$ to produce our causal estimates.
Our full experimental code is provided as an appendix.

Table \ref{tab:or_results} shows a direct comparison of our odds ratio estimates against the five separate estimation methods presented by \citet{feng2018tte}.
These numbers being less than 1 indicates our treatment reduces the probability of the outcome -- that is, TTE prevents mortality among the patient population.
Our results match those of past work, 
but produce more conservative estimates of the causal effect.
This comparison cannot prove or refute the presence of unobserved confounding via smoking status; it merely shows that when taking smoking status into account, we produce a more conservative estimate of the protective effect of TTE.


If our assumptions from \S\ref{sec: related_works} and \ref{sec: methodology} are valid, our methods should produce unbiased estimates of the causal effects.
However, with any finite sample data sample, we must worry about the variance of our methodology.
\citet{feng2018tte} also provide (Wald-type) confidence intervals for their analyses.
To produce the confidence intervals for our methods shown in Table \ref{tab:bootstrap}, we use the non-parametric bootstrap \citep{diciccio1996bootstrap}.
Bootstrapping our entire analysis from LLM pretraining to effect estimation is unfortunately prohibitively expensive; we instead use two bootstrap approaches that can give us a full picture of our method's uncertainty.

First, we consider resampling our evaluation subset of MIMIC-III to produce 100 new datasets of 4,735 patients, with each patient in each new dataset chosen independently with replacement from the original dataset.
For each of these datasets, we fit new models for $p(Y \mid X, U^*, C)$, $p(U^* \mid X, C)$, and $p(X \mid C)$.
For each of these analysis, we hold fixed the estimate for $M(U^*, U)$ computed with the n2c2 test dataset.
The 95\% confidence interval from this bootstrap alone is shown in the `MIMIC' row of Table \ref{tab:bootstrap}.


Our second, orthogonal bootstrapping method follows \citet{wooddoughty2020sensitivity}.
Rather than resampling our analysis dataset of MIMIC patients, we resample the test set of 101 patients in the n2c2 data.
For each of 100 such resamplings, we recalculate our error matrix $M(U^*, U)$, invert it to produce $I(U^*, U)$, and then use that in Equation \ref{eq:yxu_c} and (\ref{eq:uc}) to compute our causal effects. Each recalculated $M(U^*, U)$ is used alongside the original MIMIC dataset of 4,735 patients.
The 95\% confidence intervals from these 100 resamplings is shown in the `n2c2' row of Table \ref{tab:bootstrap}.

The `Both' row of the table shows a combination of both bootstrap methods.
We resample 10 new $M(U^*, U)$ matrices and 10 new MIMIC datasets on which we fit our models.
Then, for all $10\times 10 = 100$ comparisons, we compute our final estimates of the risk and odds ratios.
This gives us 100 resamplings, matching the MIMIC-only and n2c2-only bootstrap intervals.
The confidence intervals we showed previously in Table \ref{tab:or_results} use this combined bootstrap method, as it has the widest interval of the three.

Table \ref{tab:bootstrap} allows us to highlight and compare two sources of uncertainty in our method.
For the matrix adjustment method, it is unsurprising that the n2c2 bootstrap introduces more uncertainty into our final estimates, as the test set contains only 101 patients.
The matrix inverse we compute in Equation \ref{eq:deriv6} and use in (\ref{eq:yxu_c}) and (\ref{eq:uc_full}) implicitly requires us to divide by values in our estimated $M(U^*, U)$ matrix.
As with inverse propensity weighting, this can introduce high variance when dividing by small numbers \citep{ma2020robust}.
We could reduce some of this uncertainty by using a larger dataset to validate the smoking status classifier.
Interestingly, our MC-SIMEX estimates have much less variance when bootstrapping the n2c2 test set.
This may be related to the lack of theoretical foundation to explain valid estimates for standard errors in simulations \citep{lederer2006simex}.
None of our current methods are able to quantify the uncertainty introduced from the BERT pretraining or classifier fine-tuning.
While there are some methods that can capture uncertainty in neural networks (e.g., \citet{gal2016dropout}), we leave further sensitivity analyses and uncertainty quantification for future work.


\begin{table}[!t]
    \label{tab:rr_results}
    \caption{Comparison of risk ratios overall and within smoker subgroups. The left column shows our point estimates; the other columns show confidence intervals from bootstrapping either the MIMIC evaluation set or the n2c2 test set.}
    \centering
    \begin{tabular}{l ccccccc}
        & \multicolumn{3}{c}{Matrix Adjustment} & &\multicolumn{3}{c}{MC-SIMEX} \\[0.1em]
        \cmidrule{2-4} \cmidrule{6-8}
                        & RR        & MIMIC CI         & n2c2 CI & & RR        & MIMIC CI         & n2c2 CI\\
        \midrule
        Overall         & 0.88      & {[}0.85, 0.90{]} & {[}0.77, 0.99{]} & & 0.93 & {[}0.86, 1.05{]} & {[}0.93, 0.94{]} \\
        \midrule
        Past Smoker     & 0.86      & [0.76, 0.93]     & [0.32, 1.37]   & & 0.93 & {[}0.89, 0.96{]} & {[}0.93, 0.93{]} \\
        Current Smoker  & 0.92      & [0.92, 0.92]     & [0.92, 0.92]   & & 0.97 & {[}0.91, 0.97{]} & {[}0.98, 0.98{]} \\
        Non-Smoker      & 1.13      & [0.65, 1.60]     & [0.06, 2.39]   & & 0.91 & {[}0.84, 0.95{]} & {[}0.91, 0.91{]} \\
        Unknown         & 0.88      & [0.70, 1.00]     & [0.54, 1.16]   & & 0.94 & {[}0.91, 0.97{]} & {[}0.94, 0.94{]} \\
        \bottomrule
    \end{tabular}
\end{table}

Table \ref{tab:rr_results} shows our risk ratio broken down across our four subgroups of smoking statuses.
While the risk and odds ratios are closely related,
the former is \emph{collapsible}, so the risk ratio of the entire population is the weighted average of the risk ratios within each subgroup \citep{hernan2020causal}.
This allows us to easily ask whether TTE is more or less effective among patients with different smoking statuses.\footnote{In \S\ref{subsec:methods_clf} we discarded 64 patients for whom the classifier made no prediction. If we instead label those patients as ``Unknown,'' then using Matrix Adjustment and no bootstrapping, the overall risk ratio is 0.90, the odds ratio is 0.82, and the subgroup risk ratios are [0.91, 0.92, 1.15, 0.97].}
For the matrix adjustment method, for all categories except current smokers, the confidence intervals in this subgroup analysis widen and several intervals contain $1.0$, suggesting we lack the requisite data for such focused analyses.
For non-smokers the point estimate of our risk ratio is above $1.0$, which indicates that TTE is in fact harmful; while this could warrant further study, the wide intervals suggest this may be due to random variability within the small sample size.
Across both methods, an analysis of the coefficients in our $p(Y \mid X, U^*, C)$ models for the smoking status variable indicates that non-smokers have the best overall odds of survival, which aligns with past investigations of the connection between tobacco use and mortality among sepsis patients \citep{alroumi2018impact,zhang2022association}.
An unlikely but alternative explanation for the risk ratio of 1.13 could additional unobserved confounding if, hypothetically, non-smoker patients are \emph{only} administered TTE if they are at higher risk of mortality due to a factor not captured by our observed covariates $X$.

In our matrix adjustment subgroup analysis for current smokers, our methods produce a risk ratio confidence interval that has almost zero width.
This is somewhat unintuitive, especially as our MIMIC classifier predicts only 93 current smokers.
We believe this is a consequence of our test set confusion matrix from Table \ref{tab:confusion_matrix}; there are no examples in which the model falsely predicts a current smoker.
Because we do not smooth this confusion matrix, $p(U^* = \text{Current} \mid U \neq \text{Current}) = 0$ which means that our matrix adjustment represented by Equation \ref{eq:deriv6} has relatively little effect.
This helps explain the zero-width interval for the n2c2 bootstrap.
Homogeneity amongst the patients classified as current smokers could possibly explain the lack of variability from the MIMIC bootstrap.

The subgroup analyses for MC-SIMEX show an overall narrowing of confidence intervals, despite the reduced sample sizes.
As in Table \ref{tab:bootstrap}, we see very narrow intervals especially for the n2c2 bootstrap.
The average width of MC-SIMEX n2c2 bootstrap intervals is 0.05, which is rounded down to 0 with only two significant digits.
We leave for future work simulation studies and theoretical analyses that could better explore the unexpectedly low variance of this method.

\section{Conclusions and Future Work}
\label{sec: discussion}

Our method can control for unobserved confounding using an LLM classifier and a measurement error correction.
We have applied this method to estimate a causal effect with implications for clinical care, accounting for a possible violation of the assumptions of past work.
Our overall results -- that TTE protects against mortality -- comport with prior work and do not suggest any particular changes to the standards of clinical care.
However, our method more broadly offers a straightforward approach to predicting unobserved confounders from clinical notes, correcting for measurement error, and controlling for confounding.
As unobserved confounding is an untestable assumption on which all non-randomized causal analyses rely, our method demonstrates a widely-applicable method for exploring the impact of possible confounding.
For any analysis of clinical data where physicians' notes and corresponding classifiers are available, our method can be used to control unobserved confounding or as a sensitivity analysis to test the robustness of the results.

Our method relies on a number of important assumptions, and future work could explore relaxing them.
First, our choice of datasets and smoking status requires us to validate the test accuracy of our LLM classifier on the n2c2 dataset and to assume that the classifier's performance is the same on the MIMIC dataset we analyze.
In general, we would not expect the classifier to perform identically on two separate datasets, though our bootstrapping analysis suggests our results are robust to variations in our estimates of the model's performance.
This assumption could be relaxed in future work by collecting a small test set of labeled smoking statuses amongst the MIMIC patient data.
With unobserved confounders that could hypothetically be labeled by manual chart review, our method could be rendered unnecessary when sufficient time and money is available.
In practice, however, the large quantity of data involved and possible concerns about manual review of clinical notes provides strong motivation for our method.

Our approach also relies on Equation (\ref{eq:nondiff}), the assumption of non-differential error.
While it seems plausible that the classifier we use should have an error rate independent of patients' covariates, this may not be true.
In the most severe case where the error rate depends on $X, Y,$ and $C$, this would require fitting an additional model $p(U^* \mid U, X, Y, C)$.
We leave such an exploration to future work.

Our method demonstrates a general-purpose approach to adjusting for an unobserved confounder when that variable can be accurately predicted from an existing classifier with known error rate.
Extensions of this work could explore the challenges of correcting for multiple unobserved confounders with multiple classifiers.

\clearpage

\bibliography{neurips_2024}

\newpage

\section*{NeurIPS Paper Checklist}

\begin{enumerate}

\item {\bf Claims}
    \item[] Question: Do the main claims made in the abstract and introduction accurately reflect the paper's contributions and scope?
    \item[] Answer: \answerYes{} 
    \item[] Justification: We include the main claims in the abstract and introduction: proposing a methodology to take into account the measurement error caused by an unknown confounder, smoking status, and providing experiments to show the effects of measurement bias.

\item {\bf Limitations}
    \item[] Question: Does the paper discuss the limitations of the work performed by the authors?
    \item[] Answer: \answerYes{} 
    \item[] Justification: We include discussion of our assumptions and their limitations throughout the paper from \S \ref{sec: methodology} to \ref{sec: discussion}. These include the assumptions common to the matrix adjustment method and some specific to our data analysis, such as assuming our LLM classifier performs identically on the MIMIC-III data as it does on the n2c2 testing set. 

\item {\bf Theory Assumptions and Proofs}
    \item[] Question: For each theoretical result, does the paper provide the full set of assumptions and a complete (and correct) proof?
    \item[] Answer: \answerYes{} 
    \item[] Justification: We provide derivations (as well as assumptions made) on the effects of measurement error when considering an unobserved confounder when measuring causal effects in \ref{subsec:methods_me}. We also provide experiments to demonstrate the practical aspect of such derivations in \ref{sec:results}.

    \item {\bf Experimental Result Reproducibility}
    \item[] Question: Does the paper fully disclose all the information needed to reproduce the main experimental results of the paper to the extent that it affects the main claims and/or conclusions of the paper (regardless of whether the code and data are provided or not)?
    \item[] Answer: \answerYes{} 
    \item[] Justification: High-level explanations of how experiments were done are included in \ref{sec:results} as well as the derivation to implement in \ref{subsec:methods_me}. We also provide an \href{https://github.com/controllingunobservedpaper/NEURIPS_submission_2024/tree/main}{anonymized Github} that contains the implementation of the derivations in \ref{sec: methodology} and results in \ref{sec:results}. We cannot release the n2c2 or MIMIC-III datasets due to data usage agreements, but they are publicly available to anyone who accepts those agreements.

\item {\bf Open access to data and code}
    \item[] Question: Does the paper provide open access to the data and code, with sufficient instructions to faithfully reproduce the main experimental results, as described in supplemental material?
    \item[] Answer: \answerYes{} 
    \item[] Justification: We provide an \href{https://github.com/controllingunobservedpaper/NEURIPS_submission_2024/tree/main}{anonymized Github} that contains the implementation of the derivations in \ref{sec: methodology} and results in \ref{sec:results}. We cannot release the n2c2 or MIMIC-III datasets due to data usage agreements, but they are publicly available to anyone who accepts those agreements.

\item {\bf Experimental Setting/Details}
    \item[] Question: Does the paper specify all the training and test details (e.g., data splits, hyperparameters, how they were chosen, type of optimizer, etc.) necessary to understand the results?
    \item[] Answer: \answerYes{} 
    \item[] Justification: We explain our hyperparameter choices (following \citet{mulyar2019pheno}) in \S \ref{sec: methodology} and the details of our experiments (e.g., bootstrap confidence intervals, subgroup analyses) in \S \ref{sec:results}. 

\item {\bf Experiment Statistical Significance}
    \item[] Question: Does the paper report error bars suitably and correctly defined or other appropriate information about the statistical significance of the experiments?
    \item[] Answer: \answerYes{} 
    \item[] Justification: We use nonparametric bootstrap resampling to generate 95\% confidence intervals for measuring causal effects to provide additional context when explaining the effects of measurement error as well as quantify the significance our results in \ref{sec:results}. However, our main contribution is the demonstration of our causal methodology applied to a real-world dataset, not a claim of improved model performance on a given benchmark.

\item {\bf Experiments Compute Resources}
    \item[] Question: For each experiment, does the paper provide sufficient information on the computer resources (type of compute workers, memory, time of execution) needed to reproduce the experiments?
    \item[] Answer: \answerNo{} 
    \item[] Justification: The scale of experiments done in this paper do not require large numbers of CPUs and GPUs. Specifically, one 3070TI GPU was used to fine-tune the LLM and run experiments (as most relied on simple linear models) for the proposed methodology. The longest experiment (excluding model training) can be done in under an hour.
    
\item {\bf Code Of Ethics}
    \item[] Question: Does the research conducted in the paper conform, in every respect, with the NeurIPS Code of Ethics \url{https://neurips.cc/public/EthicsGuidelines}?
    \item[] Answer: \answerYes{} 
    \item[] Justification: We have adhered to the NeurIPS Code of Ethics (i.e. not releasing the n2c2 and MIMIC-III datasets) and preserved anonymity in both the paper and released code.

\item {\bf Broader Impacts}
    \item[] Question: Does the paper discuss both potential positive societal impacts and negative societal impacts of the work performed?
    \item[] Answer: \answerNA{} 
    \item[] Justification: Although the practical application of imputing smoking status as unobserved confounder and seeing the measurement bias is an important contribution, the main focus of this paper is to propose a methodology that can be used for general causal methodologies, specifically seeing the effects of measurement bias when estimating unbiased casual effects in the presence of an unobserved confounder in a pre-existing relationship between two variables.

\item {\bf Safeguards}
    \item[] Question: Does the paper describe safeguards that have been put in place for responsible release of data or models that have a high risk for misuse (e.g., pretrained language models, image generators, or scraped datasets)?
    \item[] Answer: \answerNA{} 
    \item[] Justification: Our proposed methodology does not pose such risks.

\item {\bf Licenses for existing assets}
    \item[] Question: Are the creators or original owners of assets (e.g., code, data, models), used in the paper, properly credited and are the license and terms of use explicitly mentioned and properly respected?
    \item[] Answer: \answerYes{} 
    \item[] Justification: We have included the proper citations of the specific packages utilized throughout the paper and in our list of references.

\item {\bf New Assets}
    \item[] Question: Are new assets introduced in the paper well documented and is the documentation provided alongside the assets?
    \item[] Answer: \answerNA{} 
    \item[] Justification: We do not provide new assets.

\item {\bf Crowdsourcing and Research with Human Subjects}
    \item[] Question: For crowdsourcing experiments and research with human subjects, does the paper include the full text of instructions given to participants and screenshots, if applicable, as well as details about compensation (if any)? 
    \item[] Answer: \answerNA{} 
    \item[] Justification: This paper does not involve crowdsourcing. We follow the data usage agreement for the MIMIC-III and n2c2 data.

\item {\bf Institutional Review Board (IRB) Approvals or Equivalent for Research with Human Subjects}
    \item[] Question: Does the paper describe potential risks incurred by study participants, whether such risks were disclosed to the subjects, and whether Institutional Review Board (IRB) approvals (or an equivalent approval/review based on the requirements of your country or institution) were obtained?
    \item[] Answer: \answerNA{} 
    \item[] Justification: This paper does not involve crowdsourcing. We follow the data usage agreement for the MIMIC-III and n2c2 data, and do not require additional IRB approval. 

\end{enumerate}





\end{document}